\documentclass[a4paper,fleqn]{cas-dc}
\usepackage{booktabs}
\usepackage{amsmath}
\usepackage{wrapfig}
\usepackage{subfig}
 \usepackage{subcaption}
\usepackage[accsupp]{axessibility}  
\usepackage{bm}
\usepackage{multirow}
\usepackage{tikz}
\usepackage{comment}
\usepackage{amsmath,amssymb} 
\usepackage{color}
\usepackage{pifont}
\usepackage{makecell}
\usepackage{caption}
\usepackage{float}
\usepackage{adjustbox}
\usepackage{etoolbox}
\usepackage{dsfont}

\usepackage[accsupp]{axessibility}  
\usepackage{paralist}

\usepackage{siunitx}
\usepackage{amsmath,lipsum}
\usepackage{cuted}
\usepackage{lineno}
\usepackage{booktabs,makecell,multirow}
\usepackage{amsmath,amssymb,amsfonts}
\usepackage{algorithmic}
\usepackage{graphicx}
\usepackage{textcomp}
\usepackage{amsmath,amsfonts}
\usepackage{algorithmic}
\usepackage{algorithm}
\usepackage{array}
\usepackage{textcomp}
\usepackage{stfloats}
\usepackage{url}
\usepackage{verbatim}
\usepackage{graphicx}
\usepackage{cite}
\usepackage{booktabs}
\usepackage{paralist}

\usepackage{threeparttable}

\usepackage[numbers,sort&compress]{natbib}

\def\tsc#1{\csdef{#1}{\textsc{\lowercase{#1}}\xspace}}
\tsc{WGM}
\tsc{QE}

\usepackage{xspace}
\makeatletter
\DeclareRobustCommand\onedot{\futurelet\@let@token\@onedot}
\def\@onedot{\ifx\@let@token.\else.\null\fi\xspace}

\makeatother

\usepackage{color}

\usepackage{xcolor}
\definecolor{hollywoodcerise}{rgb}{0.96, 0.0, 0.63}
\definecolor{lasallegreen}{rgb}{0.03, 0.47, 0.19}
\definecolor{hanpurple}{rgb}{0.32, 0.09, 0.98}
\definecolor{green(pigment)}{rgb}{0.0, 0.65, 0.31}

\hypersetup{colorlinks=true,linkcolor={red},citecolor={hanpurple},urlcolor={magenta}} 

\usepackage{caption}
\begin{document}
\let\WriteBookmarks\relax

\shorttitle{Towards Real-world Lens Active Alignment with Unlabeled Data via Domain Adaptation}    

\shortauthors{W.Y. Li~\textit{et al.}}  

\title [mode = title]{Towards Real-world Lens Active Alignment with Unlabeled Data via Domain Adaptation}

\author[1,5]{Wenyong Li}
\credit{Conceptualization of this study, Methodology, Software}
\author[1]{Qi Jiang}
\author[1]{Weijian Hu}[orcid=0000-0003-2053-4328]
\cormark[2]
\author[3,4]{Kailun Yang} 
\author[6]{Zhanjun Zhang}
\author[7]{Wenjun Tian}
\author[1,2]{Kaiwei Wang}[orcid=0000-0002-8272-3119]
\cormark[1]
\author[1]{Jian Bai}

%

\affiliation[a]{organization={State Key Laboratory of Extreme Photonics and Instrumentation, College of Optical Science and Engineering},
            addressline={Zhejiang University}, 
            city={Hangzhou},
            postcode={310027}, 
            country={China}}

\affiliation[b]{organization={National Research Center for Optical Instrumentation},
            addressline={Zhejiang University}, 
            city={Hangzhou},
            postcode={310027}, 
            country={China}}

\affiliation[c]{organization={School of Artificial Intelligence and Robotics},
            addressline={Hunan University}, 
            city={Changsha},
            postcode={410012}, 
            country={China}}

\affiliation[d]{organization={National Engineering Research Center of Robot Visual Perception and Control Technology},
addressline={Hunan University}, 
city={Changsha},
postcode={410082}, 
country={China}}

\affiliation[e]{organization={Intelligent Optics $\&$ Photonics Research Center, Jiaxing Research Institute},
            addressline={Zhejiang University}, 
            city={Jiaxing},
            postcode={314031}, 
            country={China}}

\affiliation[f]{organization={Dongguan Yutong Optical Technology Co., Ltd.},
            city={Dongguan},
            postcode={523863}, 
            country={China}}

\affiliation[g]{organization={Sunny Optical (Zhejiang) Research Institute Co., Ltd.},
            city={Hangzhou},
            postcode={311215}, 
            country={China}}

\cortext[1]{Corresponding author. Email: wangkaiwei@zju.edu.cn (K. Wang)}
\cortext[2]{Corresponding author. Email: huweijian@zju.edu.cn (W. Hu)}

\begin{abstract} 
Active Alignment (AA) is a key technology for the large-scale automated assembly of high-precision optical systems. Compared with labor‑intensive per‑model on‑device calibration, a digital‑twin pipeline built on optical simulation offers a substantial advantage in generating large‑scale labeled data. However, complex imaging conditions induce a domain gap between simulation and real-world images, limiting the generalization of simulation-trained models. To address this, we propose augmenting a simulation baseline with minimal unlabeled real-world images captured at random misalignment positions, mitigating the gap from a domain adaptation perspective. We introduce Domain Adaptive Active Alignment (DA3), which utilizes an autoregressive domain transformation generator and an adversarial-based feature alignment strategy to distill real-world domain information via self-supervised learning. This enables the extraction of domain-invariant image degradation features to facilitate robust misalignment prediction. Experiments on two lens types reveal that DA3 improves accuracy by $46\%$ over a purely simulation pipeline. Notably, it approaches the performance achieved with precisely labeled real-world data collected on $3$ lens samples, while reducing on-device data collection time by $98.7\%$. The results demonstrate that domain adaptation effectively endows simulation-trained models with robust real‑world performance, validating the digital‑twin pipeline as a practical solution to significantly enhance the efficiency of large-scale optical assembly.
\end{abstract}

\begin{keywords}
 \sep Active Alignment
 \sep Optical Lens Assembly
 \sep Domain Adaptation
 \sep Optical Simulation
 \sep Digital Twin
\end{keywords}
\maketitle

\section{Introduction}
\label{sec:intro}
\begin{figure*}
  \centering
  \includegraphics[width=1.0\linewidth]{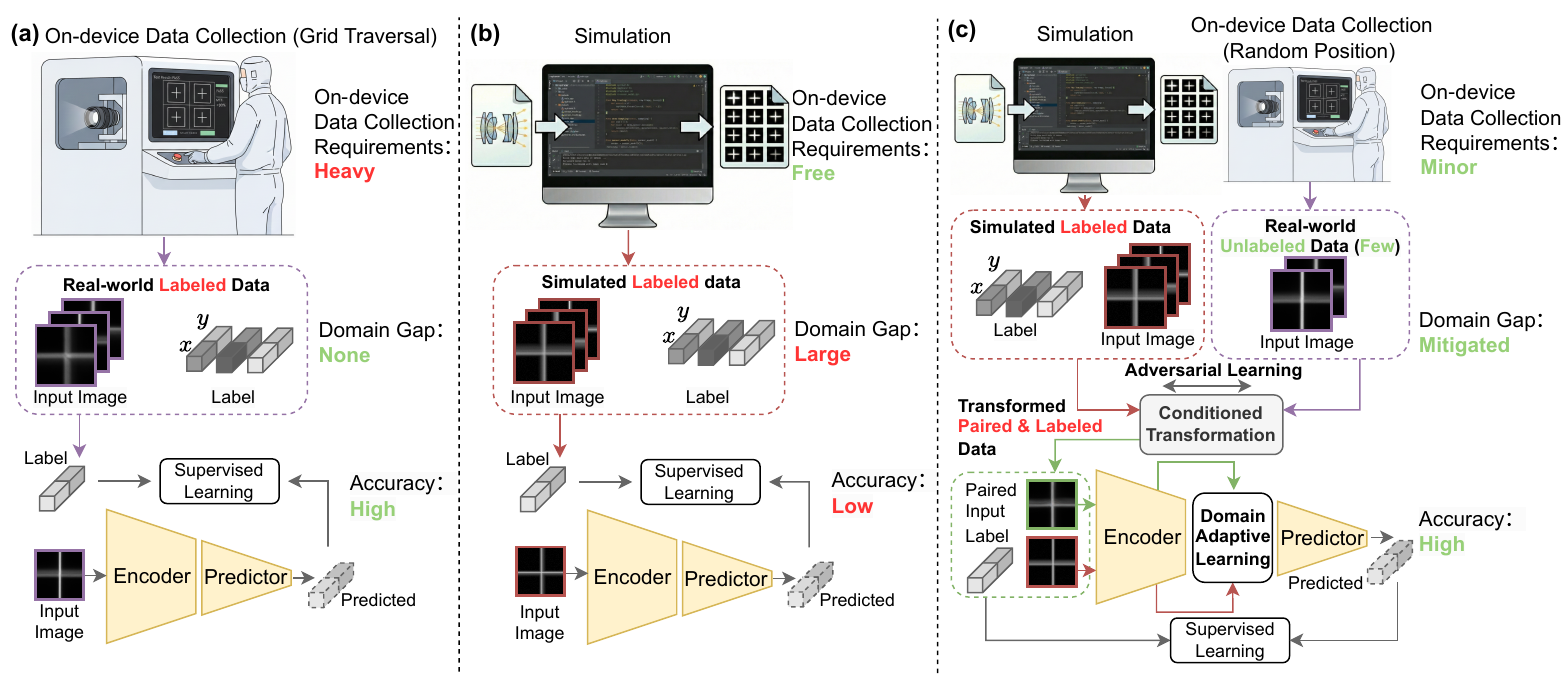}
  \caption{\textbf{Comparison of optical active alignment learning paradigms.} 
  (a) Traditional supervised learning yields high accuracy but requires high costs for on-device data collection. 
  (b) Pure simulation-based learning is free of on-device data collection yet suffers from low accuracy due to significant domain gaps. 
  (c) Our proposed framework combines pure simulation with minimal unlabeled real-world data through domain adaptation, thereby achieving high accuracy at a very low on-device data cost.
  }
  \label{fig:teaser}
\end{figure*}

Driven by the rapid escalation in sensor resolution and pixel density, the manufacturing tolerances for precision optical systems have tightened drastically, often demanding sub-micron and milliradian-level accuracy~\cite{langehanenberg2015strategies}. 
Traditional passive alignment methods, which rely solely on the geometric fit of mechanical components (\textit{e.g.}, lens barrels and spacers) to assemble elements, are increasingly insufficient~\cite{langehanenberg2015strategies,ho2017precision}. 
Due to unavoidable accumulated processing and assembly errors, these methods often fail to achieve the theoretical optical performance.
Consequently, Active Alignment (AA) has evolved from a niche technique into an indispensable process in high-end lens manufacturing~\cite{langehanenberg2015strategies,liu2024application}.

To address the efficiency and accuracy bottlenecks inherent in traditional manual adjustment, research on intelligent AA has gained significant momentum, with Deep Learning (DL) based solutions~\cite{liu2024application,burkhardt2025active,Slor:26,hu2025fast} emerging as a dominant trend. 
These methods center around training a neural network to fit the complex non-linear mapping between the imaging results and their corresponding misalignment offsets. 
However, the performance of such data-driven models is heavily contingent on the availability of large-scale and high-quality labeled image-offsets data.

Recent advances in AA data acquisition primarily fall into two paradigms, as illustrated in Fig.~\ref{fig:teaser}. 
On-device data collection (Fig.~\ref{fig:teaser}(a)) yields high prediction accuracy based on real-world labeled data with zero domain gap, but the requirement for physical manipulation makes it prohibitively time-consuming and labor-intensive.
This process involves first manually adjusting a real-world lens sample to its optimal performance position on the AA machine to establish a physical ``zero point''. 
Then, the system performs a grid traversal, mechanically shifting the lens group with fixed step sizes and ranges to capture images at known misalignment coordinates with offset labels. 
In contrast, a simulation-based digital-twin pipeline (Fig.~\ref{fig:teaser}(b)) offers a scalable route to large-scale labeled data without physical experimentation, where ray tracing and image simulation are leveraged to synthesize imaging outputs together with the corresponding misalignment offsets for the target lens under arbitrary misalignment states. 
Yet, the AA model trained on simulated data suffers from a large domain gap, making it difficult to generalize to real-world on-device validation cases, resulting in low accuracy.
The domain gap mainly stems from two distinct sources: 
(i) variations caused by diverse manufacturing tolerance combinations, which can theoretically be mitigated by introducing tolerance perturbations into the simulation; and (ii) the intrinsic imaging style gap arising from complex imaging conditions of the AA machine that are difficult to model explicitly.

To address these challenges, this paper approaches the problem from the perspective of Domain Adaptation (DA). 
Domain adaptation pipeline aims to adapt a model trained on a labeled source domain to a target domain via unsupervised or self-supervised learning, effectively utilizing unlabeled data from the target domain~\cite{wang2018deep,ganin2015unsupervised}. 
While DA strategies reveal effectiveness in high-level computer vision tasks such as object detection~\cite{chen2018domain,saito2019strong} and semantic segmentation~\cite{tsai2018learning,vu2019advent}, their potential application in simulation-based AA remains largely underexplored. 
The AA regression task shares a fundamental structural similarity with these problems, where the key lies in the extraction of robust domain-invariant features followed by a task-specific head for prediction. 

Motivated by this insight, we propose a novel simulation-driven paradigm for real-world AA. 
To establish an AA model robust to real-world variations, we leverage DA to synergize extensive labeled simulation data with a minimal set of unlabeled on-device images acquired at random misalignment positions.
The proposed Domain Adaptive Active Alignment (DA3) framework is shown in Fig.~\ref{fig:teaser}(c), where domain transformation and domain adaptive learning are integrated to distill the domain knowledge of the imaging style from the unlabeled target domain data. 
Specifically, to enable supervised learning on the target domain, we design a conditional domain transformation model to generate image-content-consistent simulated images with the target domain imaging style via adversarial learning. 
The transformed paired and labeled source domain and pseudo-target domain data are then leveraged for domain adaptive learning, where the encoder is regularized to extract domain-invariant features for robust misalignment prediction via a feature alignment strategy. 

\begin{figure}
  \centering
  \includegraphics[width=1.0\linewidth]{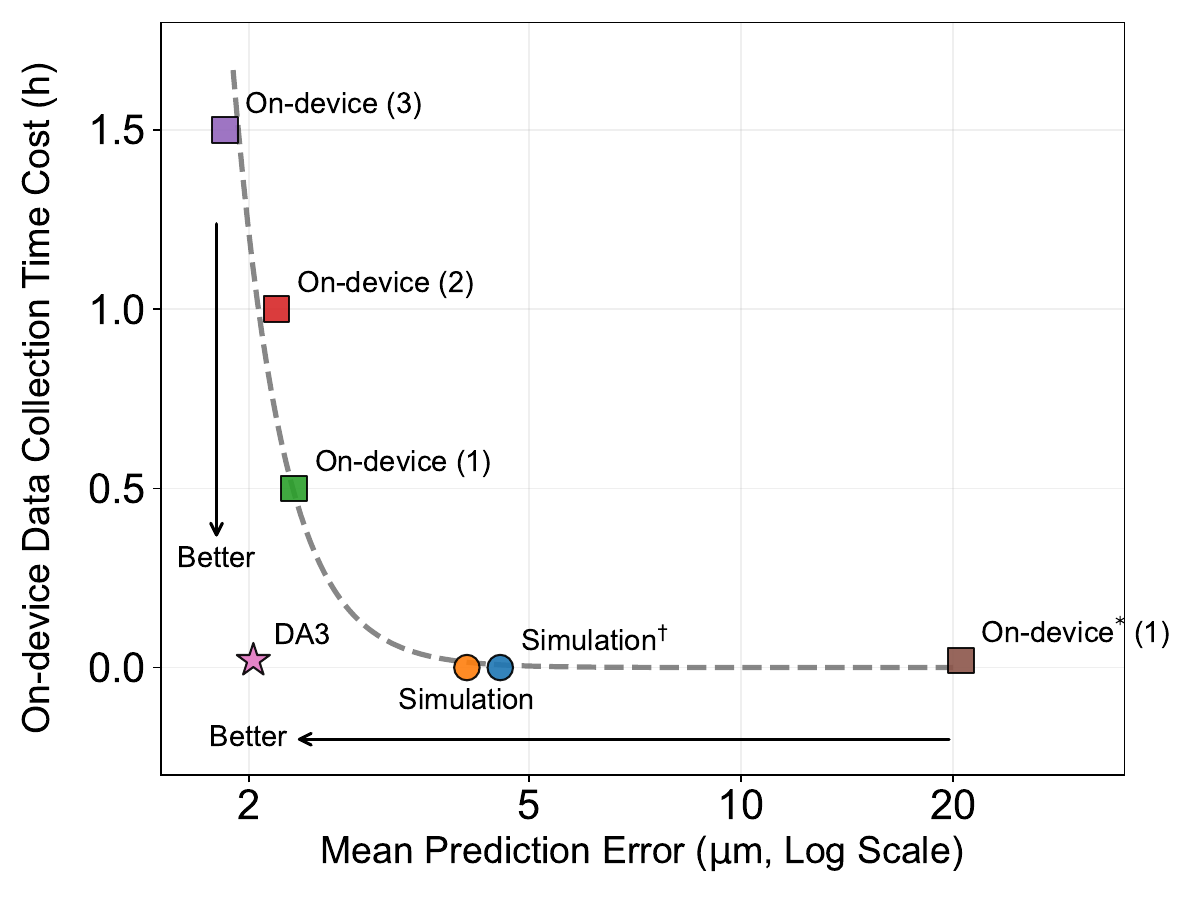}
  \caption{\textbf{Trade-off analysis between mean prediction error and data collection time cost.}
  The dashed curve illustrates the efficiency frontier of conventional supervised methods, highlighting the trade-off between accuracy and data cost.
  On-device ($N_{oracle}$) denotes training on data collected from $N_{oracle}$ real-world lenses.
  *Real-world data collected with sparse sampling. 
  $^{\dagger}$Simulation data without tolerance perturbation.}
  \label{fig:teaser2}
\end{figure}

Extensive experiments on both a real-world security lens and a simulated mobile phone lens yield $3$ key findings, with representative results for the security lens shown in Fig.~\ref{fig:teaser2}: 
(i) DA3 mitigates the prohibitive time burden of data collection while matching the precision of dense on-device benchmarks (achieving an MAE of 2.03~\si{\micro\meter});
(ii) It resolves the critical trade-off between on-device data collection requirement and prediction accuracy, whereas the model trained on sparse real-world data (10~\si{\micro\meter} step) fails, DA3 leverages only a few unlabeled on-device data to achieve high performance;
(iii) The successful bridging of the domain gap verifies the framework as a feasible foundation for digital-twin-based intelligent assembly systems.

The main contributions are summarized as follows:
\begin{compactitem}
    \item To the best of our knowledge, this work represents the first attempt to address intelligent AA from the perspective of DA, which reaches an excellent balance between data efficiency and prediction accuracy.
    \item We propose the DA3 framework, a novel paradigm for intelligent assembly that establishes a robust alignment model by synergizing data transformation with domain-invariant feature extraction. 
    \item Extensive experiments verify the efficacy of this approach, demonstrating that a model trained solely on simulation data and randomly collected unlabeled real-world images can achieve prediction accuracy comparable to benchmarks trained on labeled datasets from $3$ real-world lenses.
\end{compactitem}

\section{Related Work}
\label{sec:related}

\noindent\textbf{Active Alignment.}
Active alignment (AA) is proposed to reduce assembly errors such as decenter and tilt, thereby improving the image quality of the lens. 
Early geometry-based AA methods~\cite{langehanenberg2015strategies,ho2017precision} use high-precision measuring equipment to measure the center of curvature or the optical axis offset. However, these methods typically require complex hardware setups and suffer from low efficiency, limiting their application in large-scale production.
Consequently, performance-based methods have become widely used in the industry. These methods iteratively adjust the element position by monitoring wavefront errors or Modulation Transfer Function (MTF) values until overall optical performance is optimal. Although more practical, they rely on time-consuming search mechanisms, which remain a major bottleneck for assembly speed.
With the advent of deep learning, neural networks have been increasingly adopted to achieve "one-shot" alignment, aiming to replace iterative search.
Liu \textit{et al.}~\cite{liu2024application} employ a neural network to predict large-scale decenter errors, followed by a fine search algorithm. 
However, their supervised learning approach relies heavily on collecting large-scale labeled data from real-world devices, which is both labor-intensive and costly.
Hu \textit{et al.}~\cite{hu2025fast} propose a neural network that incorporates physical priors to estimate tolerances that conform to physical laws based on the simulated PSF. 
Burkhardt \textit{et al.}~\cite{burkhardt2025active} propose a reinforcement learning framework for AA, which learns the optimal strategy directly in the simulated sensor output pixel space to effectively address manufacturing tolerances and mechanical movement errors.
Slor \textit{et al.}~\cite{Slor:26} also propose a physics-based framework, utilizing synthetic spot diagrams and images to train deep learning models. Their approach successfully diagnoses complex 5 Degrees of Freedom (DOF) misalignment with high precision in simulated complex lens assemblies.
While these simulation-based methods avoid the cost of real-world data collection, they simultaneously introduce the critical domain gap between ideal simulations and real-world captures. Consequently, models trained solely on such synthetic data often degrade significantly when deployed on real-world AA machines.

To strike a good balance between data cost and accuracy, we propose the DA3 framework. Distinct from previous works, we address this problem from the perspective of domain adaptation. By leveraging only a minimal amount of unlabeled real-world data, DA3 effectively bridges the domain gap, enabling high-precision alignment while preserving the low-cost advantage of simulation-based pipelines.

\begin{figure}
  \centering
  \includegraphics[width=1.0\linewidth]{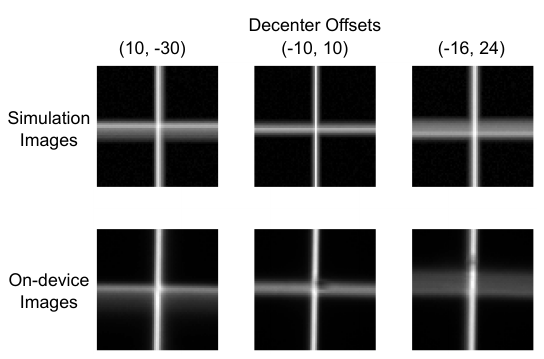}
  \caption{\textbf{Illustration of the domain gap at representative decenter offsets (Unit: \si{\micro\meter}).} 
  Although the simulation images (top) and the on-device images (bottom) exhibit similar degradations due to the same misalignment, there are distinct differences in their imaging styles.}
  \label{fig:domain_gap}
\end{figure}

\noindent\textbf{Domain Adaptation.}
Domain Adaptation (DA)~\cite{wang2018deep} aims to leverage knowledge from a labeled source domain to perform tasks in an unlabeled or sparsely labeled target domain, offering a data-efficient solution for AA where real-world data collection is prohibitive.
Mainstream DA methods can be divided into three categories, which naturally align with the specific challenges in optical assembly.

First, feature alignment methods~\cite{tzeng2014deep,long2015learning,ganin2015unsupervised} minimize distribution discrepancies via metric learning or adversarial training. This mechanism is essential for AA as it enforces the model to disregard domain-specific variations and focus on the domain-invariant optical aberration. Second, generative data transformation~\cite{ghifary2016deep,jiang2025representing} converts images into a unified style. This directly bridges the imaging style gap caused by machine-specific sensor noise and ISP pipelines, rendering simulation data indistinguishable from real captures. Third, random degradation strategies~\cite{li2024universal,li2025towards} introduce perturbations like noise or blur, effectively simulating the mechanical micro-vibrations and environmental instabilities inherent in the real-world assembly process.
DA methods are now widely applied in image classification~\cite{zhu2020deep,pandey2020target}, semantic segmentation~\cite{yang2021exploring,yue2019domain}, face recognition~\cite{sohn2017unsupervised,wang2018deep}, and other fields, but it has not yet been explored in the AA scenario.

Inspired by the above works, we introduce DA into the AA task for the first time, organically combining feature alignment, data transformation, and random degradation, aiming to achieve accurate cross-domain prediction.

\begin{figure*}
  \centering
  \includegraphics[width=1.0\linewidth]{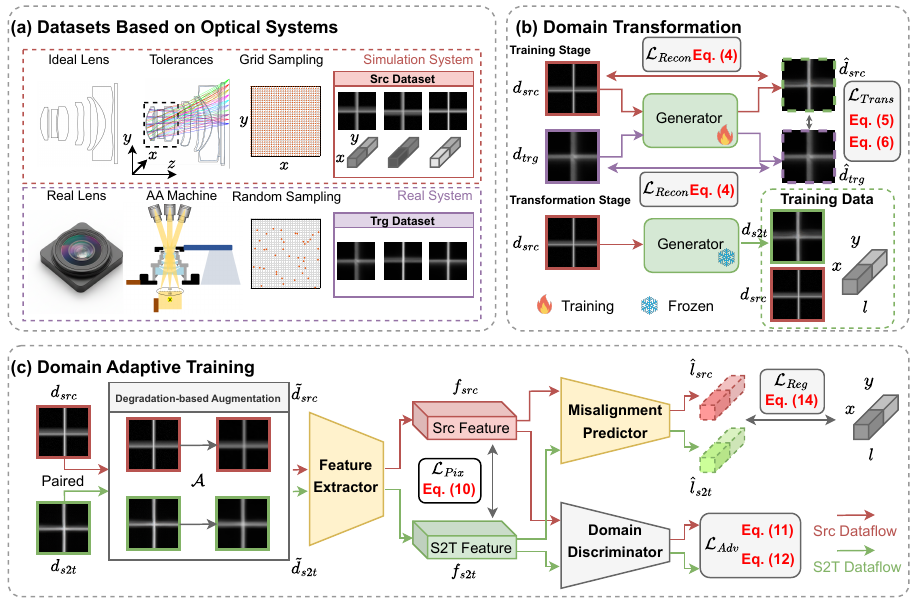}
  \caption{\textbf{Overview of the proposed domain adaptive active alignment (DA3) framework.} (a) The source dataset is constructed by grid sampling in a calibrated misalignment space through tolerance-aware optical simulation, while the target dataset is collected from several unlabeled random misalignment positions on a real-world AA machine. (b) The domain transformation module bridges the domain gap by transferring real-world imaging styles to simulated images, thereby generating content-consistent paired training data. (c) Domain adaptive training strategies achieve feature alignment across different domains by adding pixel-level consistency constraints to features extracted from degradation-based augmented paired data and performing adversarial training on them.}
  \label{fig:DA3_mainflow}
\end{figure*}

\section{Methodology}
\label{sec:method}
\subsection{Motivation}
As shown in Fig.~\ref{fig:teaser2}, existing methods based on real-world data are confronted with an unavoidable trade-off: dense sampling (On-device ($N_{oracle}$)) takes too long, whereas sparse sampling (On-device$^{*}$ ($1$)) yields poor prediction accuracy. 
While simulation provides an efficient route for data generation, models trained purely on synthetic data (Simulation$^{\dag}$) struggle to generalize to real-world cases due to the domain gap. 
Introducing tolerance perturbations (Simulation) can mitigate the gap from the perspective of the aberration characteristics of the imaging, but it still suffers from a domain gap in imaging style caused by unknown imaging conditions, resulting in only limited improvement.
As shown in Fig.~\ref{fig:domain_gap}, for the same misalignment, on-device images and simulation images exhibit similar aberration features (\textit{e.g.}, the diffusion direction and extent of the crosshair), but the diffusion in on-device data is smoother, of greater magnitude, and may exhibit spatially irregular artifacts. 
Theoretically, the mapping from aberration features to misalignment offsets remains consistent across domains, but the imaging style gap hinders accurate feature extraction from on-device data. 
Considering that on-device collection of a small number of unlabeled images at random misalignment positions is effortless, the key problem is how to leverage these target images to make the model focus only on the aberration features that reflect the misalignment offsets, while being robust to irrelevant imaging styles.

\subsection{Framework Overview}
\label{sec:overview}
In light of this, we propose the Domain Adaptive Active Alignment (DA3) framework, as illustrated in Fig.~\ref{fig:DA3_mainflow}. 
The overarching goal of DA3 is to enable high-precision alignment in the real world by effectively transferring knowledge from the simulation, utilizing only scarce unlabeled real-world data. 
As shown in Fig.~\ref{fig:DA3_mainflow}(a), the framework constructs a source dataset based on optical simulations, which endows the model with a basic ability to fit the offset and degradation features. To minimize on-device data acquisition costs, the target dataset contains only unlabeled images of a few random misalignment positions on a real-world AA machine.
To bridge the domain gap, a generative domain transformation module (Fig.~\ref{fig:DA3_mainflow}(b)) utilizes real-world imaging styles contained in the target dataset to convert the source images into pseudo-target images. Source images and pseudo-target images with the same label are paired for subsequent adversarial training.
Finally, a domain adaptive training strategy (Fig.~\ref{fig:DA3_mainflow}(c)) is employed to constrain the network to extract domain-invariant features, promoting robust prediction performance across both source and target domains.

\subsection{Data Construction for Domain Adaptation}
\label{sec:dataset}
To construct the data foundation for DA, we acquire two categories of data: extensive labeled simulation data to equip the model with the fundamental capability of misalignment regression, and a minimal set of unlabeled on-device images captured at random positions to learn domain-specific knowledge. 
The data is organized in a hierarchical structure: dataset $\rightarrow$ lens $\rightarrow$ images (\& labels).

\noindent\textbf{Tolerance-aware Optical Image Simulation.} 
To construct the source dataset $\mathcal{D}_{src}$, we establish a high-fidelity simulation model based on crosshair imaging. 
This dataset covers $1$ ideal lens and $M$ lenses with different tolerance combinations.
For each simulated lens, we employ ray tracing to calculate the Huygens PSF $h(\mathbf{v})$ at every misalignment offset $\mathbf{v}$ in the sampling grid. The simulated image $d_{src}$ is generated via convolution: 
\begin{equation}
\label{eq:ISP}
d_{src} = ISP(ISP^{-1}(d_{ideal}) \otimes h(\mathbf{v})),
\end{equation}
where $ISP$ denotes the simulation algorithm of the sensor image signal processor. 
To ensure robust training, we perform a grid sampling of the misalignment space.
The total number of samples per lens, $N_{grid}$, is determined by the scanning range and the step size. 
Mathematically, $\mathcal{D}_{src}$ is structured as a collection of lenses, where each lens contains a dense set of image-label pairs:
\begin{equation}
\label{eq:Dsrc}
\mathcal{D}_{src} = \{ L_{src}^k \}_{k=0}^{M}, \text{ where } L_{src}^k = \{ (d_{src}^i, l^i) \}_{i=1}^{N_{grid}}.
\end{equation}
Here, $L_{src}^0$ represents the ideal lens, and $L_{src}^{1...M}$ represent lenses with various assembly tolerances. 
$d_{src}$ denotes simulation images from 5 different Fields Of View (FOVs), represented by a single image in Fig.~\ref{fig:teaser} and Fig.~\ref{fig:DA3_mainflow} for brevity. $l$ is the misalignment label.

\noindent\textbf{On-device Collection of Real-world Data.}
The target dataset $\mathcal{D}_{trg}$ is collected from a specific real-world lens mounted on an AA machine.
As the target of self-supervised DA, these real-world samples do not require labels, so we can directly collect data at random positions in the decenter space without pre-alignment. 
At the same time, the total number of samples $N_{random} \ll N_{grid}$ further saves on-device data acquisition costs.
The structure of $\mathcal{D}_{trg}$ is defined as:
\begin{equation}
\label{eq:Dtrg}
\mathcal{D}_{trg} = \{ L_{trg}^1 \}, \text{where } L_{trg}^1 = \{ d_{trg}^j \}_{j=1}^{N_{random}}.
\end{equation}
Here, $d_{trg}$ also denotes images of 5 FOVs from real-world cameras.

To summarize, the data prepared for subsequent domain transformation comprises paired $\{(d_{src}, l_{src})\}$ from the source domain and unlabeled images $\{d_{trg}\}$ from the target domain.

\subsection{Domain Transformation Based on Image Generation}
\label{sec:transformation}
We construct a domain transformation model that preserves the misalignment-induced degradation characteristics present in source images, while imparting the unknown degradation style of the target domain.
Recent autoregressive generators reveal advantages in maintaining image content while injecting a global domain style, yielding strong performance in domain transformation~\cite{jiang2025representing,wang2021unsupervised}, which are well-suited for this purpose. 
Building on this insight, as shown in Fig.~\ref{fig:DA3_mainflow}, we train a domain transformation generator $G$ with unpaired source images $d_{src}$ and target images $d_{trg}$ to generate labeled source data in the target domain style, thereby enabling fully supervised learning directly in the target domain.

Concretely, given images $d_{src}$ and $d_{trg}$ randomly sampled from $\mathcal{D}_{src}$ and $\mathcal{D}_{trg}$, we employ $G$ to perform autoregressive reconstruction and supervise each reconstruction $\hat{d}_{src}=G(d_{src})$ and $\hat{d}_{trg}=G(d_{trg})$ with a pixel-wise loss function $\mathcal{L}_{Recon}$:
\begin{equation}
\label{eq:recon}
\mathcal{L}_{Recon} = {\Vert{\hat{d}_{src}-d_{src}}\Vert_1} + {\Vert{\hat{d}_{trg}-d_{trg}}\Vert_1}.
\end{equation}
Furthermore, to ensure that the outputs of $G$ comply with the target domain style, we impose an adversarial-based transformation loss $\mathcal{L}_{Trans}$ that regularizes the generation style towards the target domain:
\begin{equation}
\label{eq:trans}
\mathcal{L}^{G}_{Trans} = \mathbb{E}(D_{s2t}(\hat{d}_{src})-1)^2 + \mathbb{E}(D_{s2t}(\hat{d}_{trg})-1)^2,
\end{equation}
\begin{equation}
\label{eq:trans_d}
\begin{split}
\mathcal{L}^{D_{s2t}}_{Trans} = &\mathbb{E}(D_{s2t}(\hat{d}_{src})-0)^2
+ \mathbb{E}(D_{s2t}(\hat{d}_{trg})-0)^2  \\
+ &\mathbb{E}(D_{s2t}(d_{trg})-1)^2,
\end{split}
\end{equation}
where $D_{s2t}$ is the discriminator used to identify the target domain style; $\mathcal{L}^{G}_{Trans}$ is used to constrain $G$ to generate images that $D_{s2t}$ classifies as the target domain; and $\mathcal{L}^{D_{s2t}}_{Trans}$ constrains $D_{s2t}$ to discriminate real target domain images from the generated ones, achieving adversarial learning.

In summary, the training objective $\mathcal{L}_{s2t}$ of the domain transformation generator is the sum of $\mathcal{L}_{Recon}$ and $\mathcal{L}^{G}_{Trans}$, achieving the conversion of images toward the target domain style while keeping the image content unchanged:
\begin{equation}
\label{eq:trans_all}
\mathcal{L}_{s2t} = \mathcal{L}_{Recon} + \mathcal{L}^{G}_{Trans}.
\end{equation}
After training $G$, we translate all source domain images $d_{src}$ with $G$ to produce labeled pseudo-target domain data $\mathcal{D}_{s2t} = \{ L_{s2t}^k \}_{k=0}^{M}, \text{ where } L_{s2t}^k = \{ (d_{s2t}^i, l^i) \}_{i=1}^{N_{grid}}$:
\begin{equation}
\label{eq:s2tprocess}
d_{s2t} = G(d_{src}).
\end{equation}
The labeled translated data $\mathcal{D}_{s2t}$, alongside the original labeled source data $\mathcal{D}_{src}$, is employed for the subsequent domain adaptive training.

\subsection{Domain Adaptive Training Based on Feature Alignment}
\label{sec:DANN}
Since we focus on the training framework, models of any structure can be applied in principle.
Without loss of generality, we follow the previous DL-based AA method~\cite{liu2024application} and employ a streamlined architecture as shown in Fig.~\ref{fig:DA3_mainflow}(c), where a CNN-based feature extractor $E$ is applied to extract aberration-aware features $f = E(\tilde{d})$ from the input image $d$ for predicting the corresponding misalignment offsets $\hat{l} = P(f)$ based on an MLP-based predictor $P$.

The proposed DA training centers around extracting domain-invariant features. 
We aim to regularize $E$ to filter out domain-specific style information (\textit{e.g.}, sensor noise style) and focus solely on the aberration-aware diffusion of the crosshair caused by misalignment. 
Provided that $E$ can extract consistent aberration-aware representations from cross-domain images with identical misalignment, the predictor $P$ is enabled to perform domain-robust prediction.
To this intent, the network is simultaneously fed pairs of source data $d_{src}$ and generated pseudo-target data $d_{s2t}$ under the same misalignment offsets for the following DA training. 

\noindent\textbf{Degradation-based augmentation.} 
Although $d_{s2t}$ helps bridge the domain gap, it essentially represents a generated intermediate domain. 
To mitigate overfitting to this intermediate domain, we propose a stochastic degradation-based augmentation strategy. 
By exposing the model to diverse domain shifts at each training iteration, this approach compels the network to focus on learning domain-invariant representations rather than memorizing domain-specific styles.
The augmentation module $\mathcal{A}$ is implemented as a unified stochastic mapping function. 
For an input image $d$, the augmented output $\tilde{d}$ is defined as:
\begin{equation}
\label{eq:trans_d_}
\tilde{d} = \mathcal{A}(d; \tau, \theta) = \mathcal{T}_{\tau}(d; \theta), \quad \text{where } \mathcal{T}_{\tau} \in \mathbb{D}.
\end{equation}
Here, $\mathbb{D}$ represents the set of candidate degradations, $\tau$ denotes the specific degradation type randomly sampled from the set, and $\theta$ represents the corresponding degradation parameters. Inspired by the degradation space proposed in BSRGAN~\cite{liu2022blind}, our candidate set $\mathbb{D}$ includes JPEG compression, Gaussian blur, Gaussian noise, and random masking.

\noindent\textbf{Paired feature consistency.} 
Since the content-aligned augmented data $\tilde{d}_{src}$ and $\tilde{d}_{s2t}$ share the identical misalignment label, we employ pixel-wise constraints to encourage the shared $E$ to extract the same features from images of diverse imaging styles, aiming for consistent misalignment prediction across different domains. 
This is achieved via a pixel-wise loss $\mathcal{L}_{Pix}$ between the extracted features $f_{src} = E(\tilde{d}_{src})$ and $f_{s2t} = E(\tilde{d}_{s2t})$:
\begin{equation}
\label{eq:Lpix}
\mathcal{L}_{Pix} = || f_{src} - f_{s2t} ||_1.
\end{equation}

\noindent\textbf{Adversarial feature alignment.} 
While $\mathcal{L}_{Pix}$ provides pixel-wise feature alignment, we further employ adversarial learning to ensure the overall feature distributions of the source domain and the pseudo-target domain are indistinguishable. 
The discriminator $D_{d}$ aims to classify the source domain as the true domain and the pseudo-target domain as 0, while $E$ aims to fool it. Following the standard DA paradigm~\cite{ganin2016domain}, we employ the binary cross-entropy objective for the adversial loss $\mathcal{L}^{D_{d}}_{Adv}$ and $\mathcal{L}^{E}_{Adv}$: Discriminator loss $\mathcal{L}^{D_{d}}_{Adv}$ trains $D_{d}$ to distinguish between the extracted features $f_{src} = E(\tilde{d}_{src})$ and $f_{s2t} = E(\tilde{d}_{s2t})$:
\begin{equation}
\label{eq:LD}
\mathcal{L}^{D_{d}}_{Adv} = -\mathbb{E}[\log(D_{d}(f_{src}))] - \mathbb{E}[\log(1 - D_{d}(f_{s2t}))].
\end{equation}
Conversely, $E$ aims to force the outputs of $D_{d}$ to approach the decision boundary of 0.5:
\begin{equation}
\begin{split}
\label{eq:Ladv}
\mathcal{L}^{E}_{Adv} = & - \frac{1}{2} \mathbb{E} \left[ \log \Big( D_{d}(f_{src}) \cdot (1 - D_{d}(f_{src})) \Big) \right]\\
& - \frac{1}{2} \mathbb{E} \left[ \log \Big( D_{d}(f_{s2t}) \cdot (1 - D_{d}(f_{s2t})) \Big) \right].
\end{split}
\end{equation}

\noindent\textbf{End-to-end training objective.} 
Based on the above feature alignment constraints, the entire framework is trained in an end-to-end manner via $\mathcal{L}_{total}$, integrating the primary regression task on the misalignment offsets:
\begin{equation}
\label{eq:Ltotal}
\mathcal{L}_{total} = \mathcal{L}_{Reg} + \lambda_{pix}\mathcal{L}_{Pix} + \lambda_{adv}\mathcal{L}^{E}_{Adv},
\end{equation}
where
\begin{equation}
\label{eq:Lreg}
\mathcal{L}_{Reg} = || P(f_{src}) - l ||_2^2 + || P(f_{s2t}) - l ||_2^2
\end{equation}
is the misalignment prediction loss calculated on $f_{src}$ or $f_{s2t}$, and $\lambda_{pix}, \lambda_{adv}$ are hyperparameters balancing the contributions of the alignment terms.

\section{Experiments and Results}
\label{sec:exp}
\subsection{Implementation Details}
\label{exp:details}
\begin{figure}
  \centering
  \includegraphics[width=1.0\linewidth]{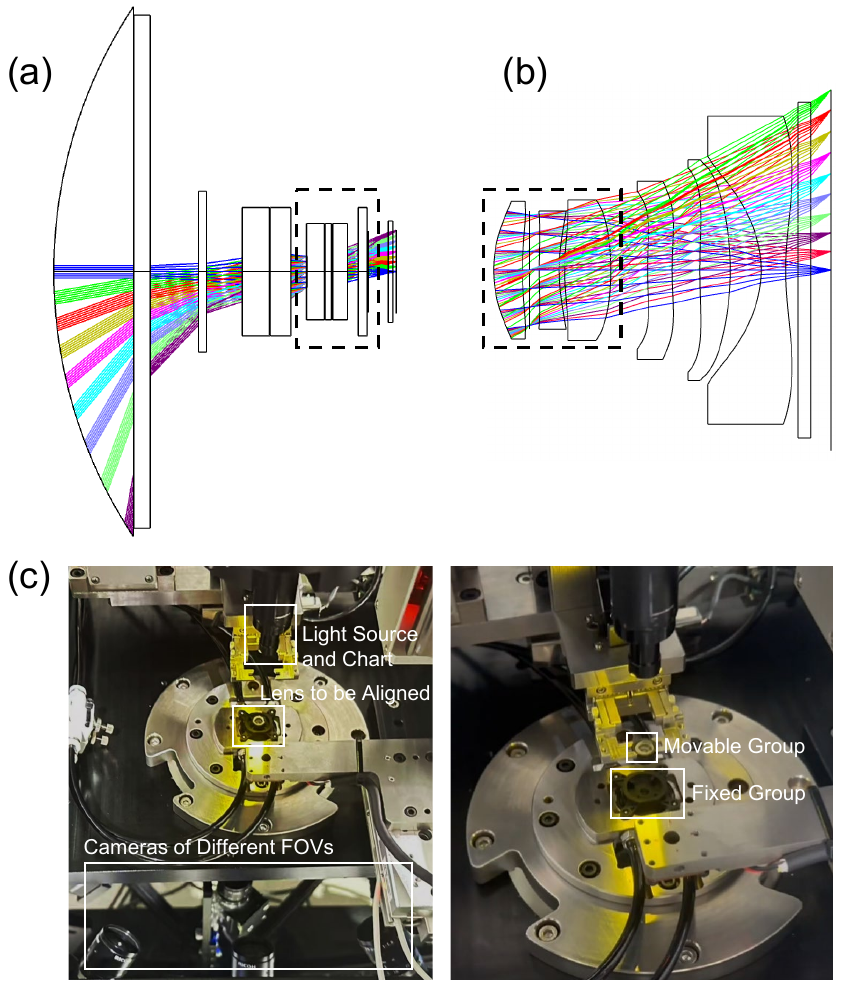}
  \caption{\textbf{Optical structures of the experimental lenses and the real-world AA machine.} 
  (a) The black-box model of the security lens.
  (b) The 7P aspheric smartphone lens.
  In both diagrams, the lens groups enclosed in dashed rectangles indicate the movable components adjusted during the AA process, while the remaining elements are fixed.
  (c) The real-world AA machine setup utilized for on-device data acquisition and lens assembly.}
  \label{fig:lens&machine}
\end{figure}

\noindent\textbf{Evaluation Protocol.}
To rigorously evaluate the proposed DA3 model, we establish a standardized testing protocol where multiple real-world lenses corresponding to the simulated design serve as the test set. As our experimental validation is conducted solely on decenter, we refer to misalignment as decenter in the following sections.
To construct the test set, we implement a two-stage procedure on the AA machine (Fig.~\ref{fig:lens&machine}(c)): pre-alignment and data acquisition. 
In the pre-alignment phase, we perform a large-scale grid scan for each lens to identify the position with the highest multi-field weighted MTF value, marking it as the origin. 
Subsequently, we perform a high-precision grid scan centered on this origin to collect decenter samples, each containing images taken from 5 different FOVs. 
During evaluation, the model uses these 5-FOV images as a single input to predict the decenter offset. 
We employ the Mean Absolute Error (MAE) and Standard Deviation (SD) of the prediction results as the primary metrics to quantify alignment accuracy and stability.

\noindent\textbf{Data Collection.}
We utilize two distinct experimental settings to validate our method: a mass-produced security lens setting using real-world data, and a 7P aspherical smartphone lens setting using simulated data to model a controllable domain gap, as illustrated in Fig.~\ref{fig:lens&machine}. 
Tab.~\ref{tab:settings} outlines the specific parameters and differences between these two configurations.

\begin{table}[h!]
    \begin{center}
        \caption{\textbf{Experimental settings and specifications.}The specifications cover the data generation methods, dataset splits, and the sampling parameters for both the security lens and smartphone lens.}
        \label{tab:settings}
        \resizebox{0.5\textwidth}{!}{
\renewcommand{\arraystretch}{1.25}
\setlength{\tabcolsep}{1mm}
\begin{tabular}{@{}ccccll@{}}
\toprule[0.17em]
\textbf{Specification} & \textbf{Security Lens} & \textbf{Smartphone Lens} \\
\midrule
Source Data Origin & Simulation(w. ISP) &  Simulation(w.o. ISP) \\
Target Data Origin & Real-world Capture &  Simulation(w. ISP) \\
Train Set Size & 11 Lenses & 11 Lenses \\
Test Set Size & 17 Lenses & 20 Lenses \\
Oracle Train Set Size& 3 Lenses & 3 Lenses \\
Input Image Size & $\text{70} \times \text{70}$ pixels & $\text{50} \times \text{50}$ pixels \\
Decenter Range & $\pm$ 30 \si{\micro\meter} & $\pm$ 15 \si{\micro\meter} \\
Step Size & 2 \si{\micro\meter} & 1 \si{\micro\meter} \\ \bottomrule
\end{tabular}
}

    \end{center}
\end{table}

For the simulation data, we employ the ZOS-API to generate PSFs and convolve them with an ideal crosshair pattern, followed by ISP simulation to introduce realistic noise and gamma correction. For the real-world security lens data, images are captured, cropped, and downsampled to match the simulated input dimensions.
Detailed dataset statistics are provided in Tab.~\ref{tab:settings}. 
We denote the number of lenses in the source training set, the target test set, and the target oracle set as $N_{train}$, $N_{test}$, and $N_{oracle}$, respectively. The source domain consists of $N_{train} = 11$ simulated lenses (1 ideal and 10 with tolerances). 
The target domain data is partitioned into two mutually exclusive subsets: the primary test set containing $N_{test}$ lenses (17 for security, 20 for smartphone) used strictly for evaluation, and a separate oracle set containing $N_{oracle} = 3$ additional lenses used exclusively to train the supervised upper-bound baseline.
Detailed procedures for data generation and preprocessing are provided in the supplementary material.

\noindent\textbf{Network Architecture.}
For the domain transformation generator $G$, we adopt the VQGAN-based structure from~\cite{jiang2025representing}, which utilizes a codebook-based bottleneck to effectively translate aberration styles. 
For the alignment network, we employ a ResNet-18~\cite{he2016deep} backbone pre-trained on ImageNet as the feature extractor $E$. The first convolutional layer is modified to accept concatenated tensors of shape $[B, N \times C, H, W]$ to handle multi-view inputs. The extracted 512-dimensional features are fed into two parallel Multi-Layer Perceptron (MLP) branches: a label predictor $P$ for regression and a domain classifier $D$ for domain discrimination.
Specific architectural details for all modules are detailed in the supplementary material.

\noindent\textbf{Training Details.}
We implement our method using PyTorch on a single NVIDIA GeForce RTX 3090.
The AA model under the DA3 framework is trained for $45K$ iterations with a batch size of 64 using the Adam optimizer. We employ a differential learning rate strategy to stabilize the adversarial training: the learning rate is set to $1 \times 10^{-3}$ for $E$ and $P$, while $D$ uses a lower learning rate of $1 \times 10^{-4}$.
To bridge the domain gap and improve robustness, we apply Gaussian blur as the exclusive degradation-based augmentation strategy, based on the ablation results in Sec.~\ref{exp:ab}.
Regarding the loss function hyperparameters, we empirically determine the optimal values to be $\lambda_{adv} = 1$ for the adversarial loss and $\lambda_{pix} = 0.05$ for the feature consistency loss.

More details can be found in the supplemental material.

\subsection{Comparison with Previous AA Pipelines}
\begin{table*}[h!]
\centering
\caption{\textbf{Quantitative comparison with baseline AA pipelines on two lens cases. (Unit: \si{\micro\meter}).}
The metrics reported are Mean Absolute Error (MAE) and Standard Deviation (SD). On-device ($N_{oracle}$) denotes training on real-world data collected from $N_{oracle}=1, 2, 3$ lenses with dense sampling steps, serving as the reference upper bound. On-device$^*$ indicates training on real-world data with large sampling steps to reduce collection costs. 
The symbol $\dagger$ represents methods trained without simulating lens tolerance perturbations. DA3 denotes our proposed domain adaptation framework. 
Best results among the non-oracle methods are highlighted in bold.}
\label{tab:main}

\resizebox{1\textwidth}{!}{
\renewcommand{\arraystretch}{1.25}
\setlength{\tabcolsep}{1mm}
\begin{tabular}{lcccccc}
\toprule[0.17em]
\multirow{2}{*}{\textbf{Pipeline}} &
\multicolumn{3}{c}{\textbf{Case 1: Security Lens}} &
\multicolumn{3}{c}{\textbf{Case 2: Smartphone Lens}} \\
\cmidrule{2-7}
& 
\textbf{MAE$_{\text{X}}$ $\pm$ SD$_{\text{X}}$} &
\textbf{MAE$_{\text{Y}}$ $\pm$ SD$_{\text{Y}}$} &
\textbf{MAE$_{\text{Avg}}$ $\pm$ SD$_{\text{Avg}}$} &
\textbf{MAE$_{\text{X}}$ $\pm$ SD$_{\text{X}}$} &
\textbf{MAE$_{\text{Y}}$ $\pm$ SD$_{\text{Y}}$} &
\textbf{MAE$_{\text{Avg}}$ $\pm$ SD$_{\text{Avg}}$} \\
\cmidrule{1-7}
On-device (1) & 
\text{2.40 $\pm$ 1.99} & \text{2.24 $\pm$ 1.73} & \text{2.32 $\pm$ 1.87} & \text{7.09 $\pm$ 5.08} & \text{6.53 $\pm$ 4.64} & \text{6.81 $\pm$ 4.87}  \\

On-device (2) & \text{2.52 $\pm$ 2.01} & \text{1.86 $\pm$ 1.45} & \text{2.19 $\pm$ 1.79} & \text{4.43 $\pm$ 4.26} & \text{6.04 $\pm$ 4.91} & \text{5.24 $\pm$ 4.67} \\

On-device (3) & \text{1.55 $\pm$ 1.30} & \text{2.15 $\pm$ 1.76} & \text{1.85 $\pm$ 1.58} & \text{4.46 $\pm$ 5.00} & \text{5.56 $\pm$ 4.82} & \text{5.01 $\pm$ 4.94} \\

On-device$^{*}$ (1) & \text{17.01 $\pm$ 10.63} & \text{24.04 $\pm$ 14.36} & \text{20.52 $\pm$ 13.11} & \text{12.94 $\pm$ 9.01} & \text{7.83 $\pm$ 4.76} & \text{10.83 $\pm$ 7.64}  \\

Simulation$^{\dag}$ & \text{3.04 $\pm$ 2.63} & \text{6.07 $\pm$ 4.75} & \text{4.55 $\pm$ 4.13} & \text{7.66 $\pm$ 6.66} & \text{7.95 $\pm$ 4.99} & \text{7.81 $\pm$ 5.89} \\

Simulation  & \text{3.36 $\pm$ 4.31} & \text{4.80 $\pm$ 4.87} & \text{4.08 $\pm$ 4.65} & \text{6.52 $\pm$ 4.74} & \text{7.23 $\pm$ 4.59} & \text{6.88 $\pm$ 4.68} \\
\cmidrule{1-7}

DA3$^{\dag}$ & \text{2.37 $\pm$ 2.03} & \text{2.82 $\pm$ 2.21} & \text{2.60 $\pm$ 2.13} & \text{4.15 $\pm$ 3.67} & \text{4.23 $\pm$ 3.23} & \text{4.19 $\pm$ 3.46} \\

DA3 & \textbf{2.03 $\pm$ 2.02} & \textbf{2.03 $\pm$ 1.80} & \textbf{2.03 $\pm$ 1.92} & \textbf{3.97 $\pm$ 3.96} & \textbf{3.71 $\pm$ 4.18} & \textbf{3.84 $\pm$ 4.07} \\
\bottomrule
\end{tabular}
}
\end{table*}
Tab.~\ref{tab:main} presents the quantitative results of our proposed DA3 framework compared against various baseline AA pipelines. 
To establish a rigorous performance upper bound, we introduce the On-device ($N_{oracle}=1,2,3$) setting, where the model is trained on the full oracle set consisting of the 3 real-world lenses detailed in Sec.~\ref{exp:details}. 
We also define On-device* (1), a low-cost baseline trained on a single oracle lens using a $5\times$ larger sampling step size. 
The symbol $^{\dagger}$ denotes methods trained without simulating lens tolerance perturbations. 

\noindent\textbf{Impact of sampling density and simulation.} 
Comparing the results of rows 1, 4, 5, and 7 reveals the critical role of data
 sampling density. While On-device (1) achieves low MAE using dense real-world data, reducing the sampling density to save time in On-device* (1) causes a severe performance drop (\textit{e.g.}, MAE surges from 2.32 to 20.52 \si{\micro\meter} in security lens). 
This indicates that sparse sampling of real-world data fails to provide sufficient degradation features for the model to learn. However, Simulation$^\dagger$ utilizes dense synthetic data to compensate for the missing information caused by sparse sampling, significantly outperforming the On-device* (1) baseline. 
Furthermore, when combined with DA, the performance of DA3$^\dagger$ improves further, approaching the level of the dense real-world baseline On-device (1).

\noindent\textbf{Impact of tolerance perturbations and domain adaptation.} 
An analysis of rows 5 through 8 isolates the contributions of tolerance perturbations and DA. 
First, comparing the methods with and without tolerance perturbations (Simulation$^{\dagger}$ \textit{vs.} Simulation, and DA3$^{\dagger}$ \textit{vs.} DA3) demonstrates that introducing tolerance perturbations consistently reduces prediction error (\textit{e.g.}, improving MAE from 4.55 to 4.08 \si{\micro\meter} in the security lens). 
This confirms that modeling manufacturing variations enhances the model's generalization across different lens samples. However, generalization via tolerance alone is insufficient, as the Simulation baseline (row 6) remains limited by the domain gap. By further incorporating DA, DA3 (row 8) significantly lowers the MAE to 2.03 \si{\micro\meter}. 
This indicates that while tolerance perturbations improve generalization, DA is essential for bridging the domain gap, ultimately enabling accurate cross-domain prediction.

\subsection{Analysis of AA Error Distribution}
\begin{figure}
  \centering
  \includegraphics[width=1.0\linewidth]{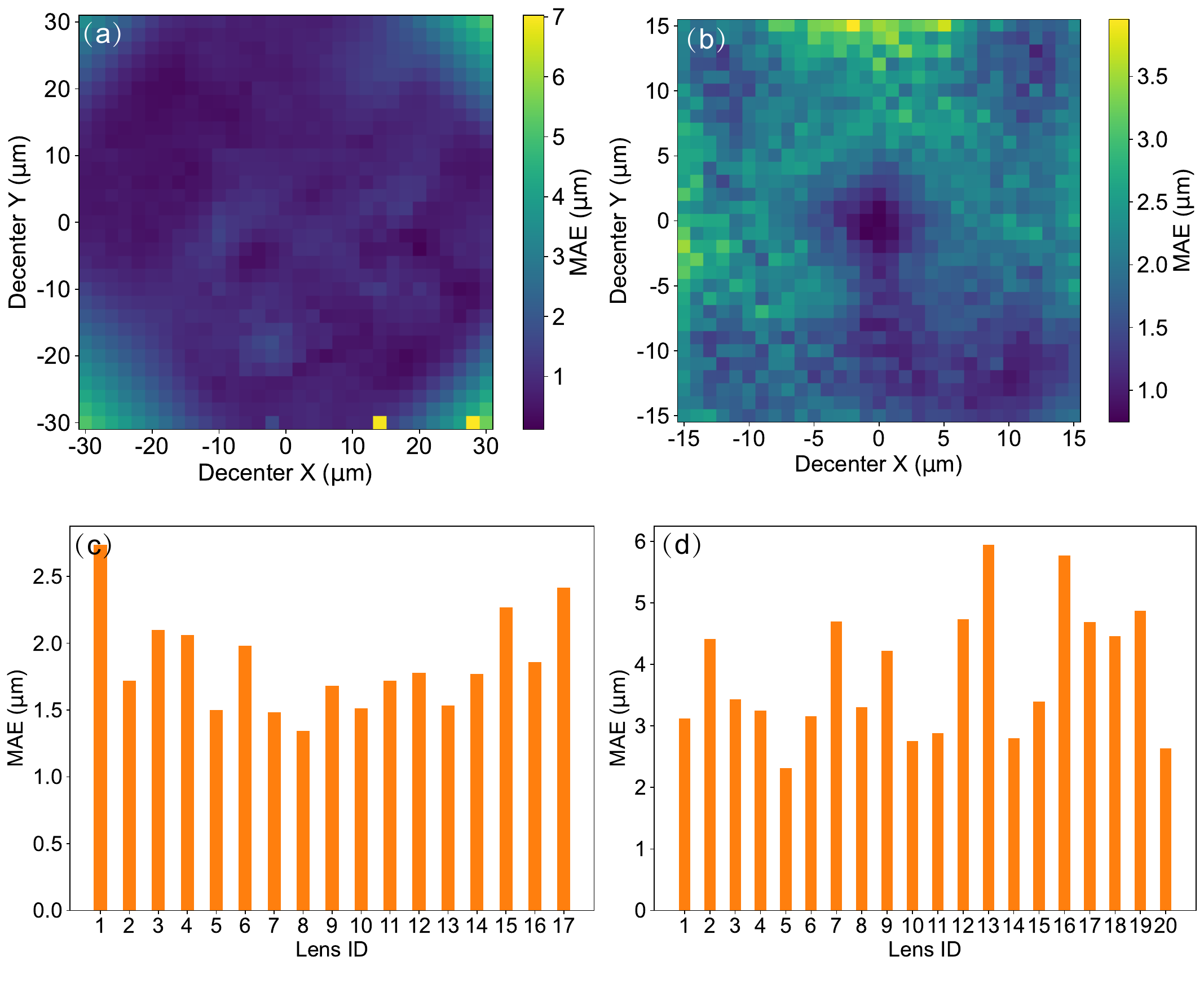}
  \caption{\textbf{Analysis of error distribution.} 
  Heatmaps (a) \& (b) visualizing the MAE distribution across the full range of decenter space for the security lens and smartphone lens. 
  Brighter colors indicate higher errors. 
  Bar charts (c) \& (d) showing the average MAE for each individual test lens in the security lens dataset (17 lenses) and smartphone lens dataset (20 lenses).}
  \label{fig:mae_per_label&lens}
\end{figure}
We further analyze the MAE distribution of our DA3 method across different decenter positions and lens instances, as illustrated in Fig.~\ref{fig:mae_per_label&lens}. 
Fig.~\ref{fig:mae_per_label&lens}(a) and~\ref{fig:mae_per_label&lens}(b) display the spatial distribution of the MAE across the two-dimensional decenter space for the security lens ($\pm$ 30 \si{\micro\meter} along both $x$ and $y$ axes) and the smartphone lens ($\pm$ 15 \si{\micro\meter} along both axes), respectively.

A consistent pattern emerges in both cases: the error is lowest represented by dark blue in the central region and gradually increases towards the boundaries. This phenomenon aligns with optical principles, as optical degradation (such as coma and astigmatism) becomes significantly more severe and non-linear at large decenter positions. This severe degradation attenuates high-frequency signal components, making the feature extraction process highly susceptible to real-world noise, thus reducing prediction accuracy. We observe a few distinct outliers represented by bright yellow spots at the extreme edges of the decenter space. 
These are attributed to cases where severe degradation causes a near-total loss of high-frequency information, making it difficult for the network to regress precise coordinates. 
However, despite the reduced precision at these extreme positions, the predicted adjustment direction remains correct, ensuring the lens is effectively guided into the central high-precision zone for final alignment.

Fig.~\ref{fig:mae_per_label&lens}(c) and \ref{fig:mae_per_label&lens}(d) present the average MAE for each individual lens in the test sets. For the security lens, the MAE across 17 real-world lenses fluctuates stably between 1.5 \si{\micro\meter} and 2.8 \si{\micro\meter}. 
Similarly, for the smartphone lens, the error across 20 simulated test lenses remains within a reasonable range without exhibiting extreme anomalies. 
The absence of outliers among the test lenses indicates two key points: first, the collected dataset is of high quality and does not contain samples with physical defects; second, our model possesses strong generalization capabilities, effectively handling the tolerance variations inherent in different lens instances.

\subsection{Visualization of the AA Process}
\begin{figure*}
  \centering
  \includegraphics[width=1.0\linewidth]{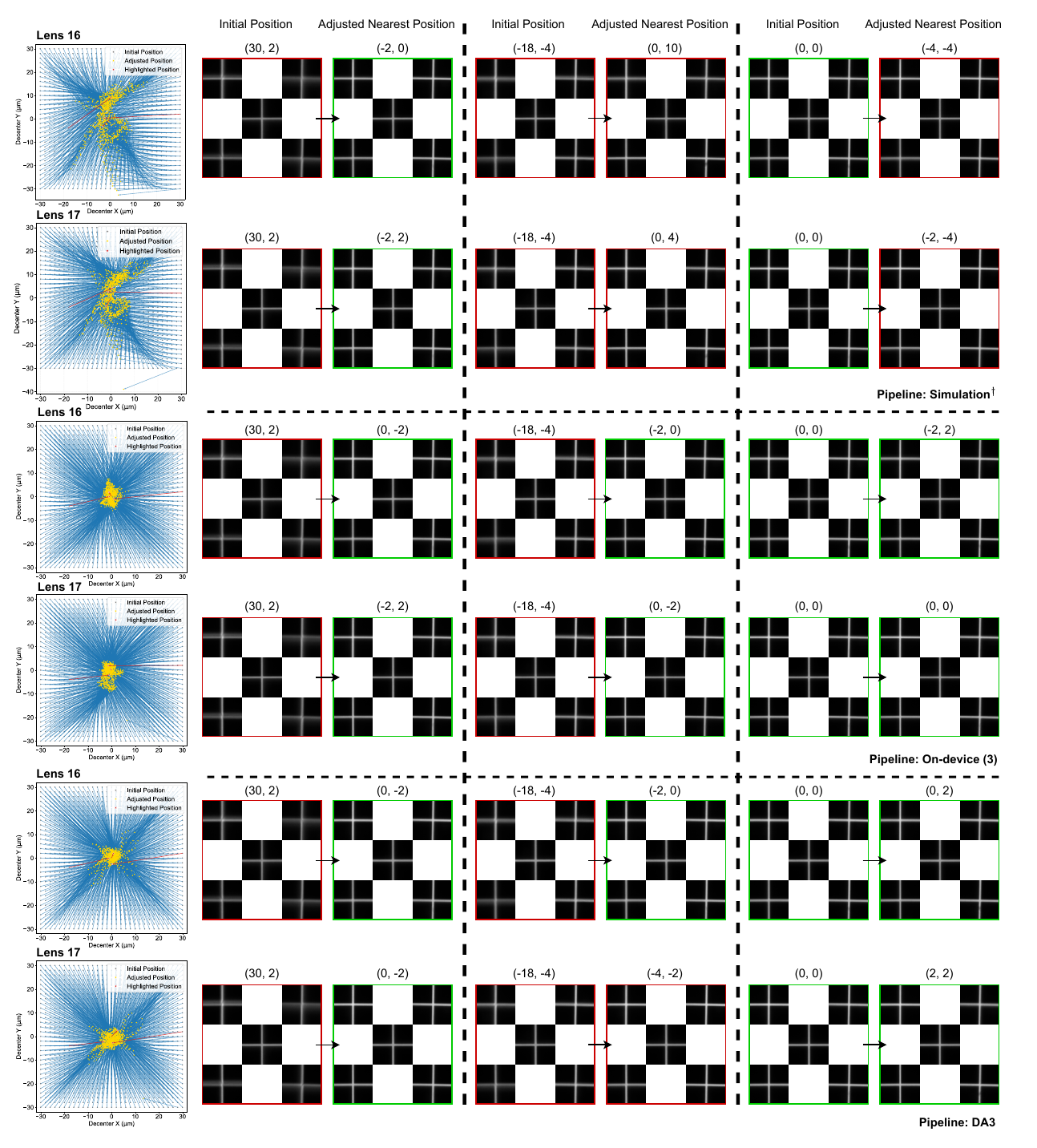}
  \caption{\textbf{Visualization of the adjustment process on two test security lenses (No. 16 and No. 17).} 
  The leftmost column plots the adjustment trajectories (blue arrows) and final positions (yellow dots) for all test cases. 
  The subsequent columns display the specific 5-FOV images after adjustment (selected via nearest neighbor) from three representative initial positions: large offset, medium offset, and zero offset. Green boxes indicate successful alignment (residual error $\leq$ 2 \si{\micro\meter} in both $x$ and $y$), whereas red boxes indicate failure.}
  \label{fig:single_test}
\end{figure*}

To intuitively evaluate the effectiveness of our method, we visualize the single-step adjustment results on two representative real-world lenses (Lens 16 and Lens 17) in Fig.~\ref{fig:single_test}.
The leftmost column displays the spatial distribution of the adjusted positions (yellow dots) relative to the optical center after one inference step for all sampled initial positions. The subsequent six columns present specific imaging examples, showing the pre- and post-adjustment comparisons for three representative initial decenter levels: large, medium, and small. A green box indicates a successful alignment (residual error $< 2$ \si{\micro\meter} in both $x$ and $y$ axes), while a red box denotes failure.
Analyzing the Simulation$^\dagger$ baseline shown in the top 2 rows, the overall performance is unsatisfactory. The scatter plots in the first column reveal that the adjusted positions fail to converge effectively toward the center, remaining widely dispersed.
Furthermore, the model exhibits severe stability issues in the small decenter offset. As shown in the two rightmost columns, the model predicts a large offset for an already aligned lens, causing drift and resulting in a failure. 
This suggests that the domain gap prevents the simulation-based model from learning reliable features, leading to performance degradation in most conditions.
In contrast, the bottom 2 rows corresponding to our DA3 framework demonstrate robust alignment capabilities comparable to the On-device oracle (middle 2 rows). 
The scatter plot shows that the adjusted positions are closely clustered near the origin. 
The image examples further verify that the DA3 not only effectively corrects large and medium offsets, but also maintains high stability for small offsets, accurately keeping the lens within the acceptable range without causing reverse optimization.

\subsection{Ablation Study}
\label{exp:ab}
Without loss of generality, we conduct the ablation study exclusively on the security lens dataset to investigate the contribution of individual components within our proposed framework. 
As the experimental trends are consistent across different lens types, this setting serves as a representative benchmark for analyzing the effectiveness of our design choices.

\begin{table}[h!]
    \begin{center}
        \caption{\textbf{Ablation study on data flow configurations (Unit: \si{\micro\meter}).} $D_{src}^{\dagger}$ and $D_{src}$ denote simulation data generated from the ideal lens configuration and the tolerance-aware configuration, respectively. The $D_{s2t}$ terms indicate the domain-transformed data generated by the corresponding domain transformation generator $G$.}
        \label{tab:ab_tol&s2t}
        \resizebox{0.5\textwidth}{!}{
\renewcommand{\arraystretch}{1.25}
\setlength{\tabcolsep}{1mm}
\begin{tabular}{@{}ccccc@{}}
\toprule[0.17em]
\textbf{Training Data} & \textbf{Type of $G$} &
\textbf{MAE$_{\text{X}}$ $\pm$ SD$_{\text{X}}$} &
\textbf{MAE$_{\text{Y}}$ $\pm$ SD$_{\text{Y}}$} &
\textbf{MAE$_{\text{Avg}}$ $\pm$ SD$_{\text{Avg}}$} \\ \midrule
$D^{\dagger}_{src}$  & \multirow{2}{*}{-}  & \text{3.04 $\pm$ 2.63} & \text{6.07 $\pm$ 4.75} & \text{4.55 $\pm$ 4.13} \\
$D_{src}$ & & \text{3.36 $\pm$ 4.31} & \text{4.80 $\pm$ 4.87} & \text{4.08 $\pm$ 4.65} \\
$D^{\dagger}_{src}+D^{\dagger}_{s2t}$ & \multirow{2}{*}{CycleGAN}  & \text{5.75 $\pm$ 4.09} & \text{3.18 $\pm$ 2.46} & \text{4.47 $\pm$ 3.61}     \\
$D_{src}+D_{s2t}$ & & \text{4.36 $\pm$ 3.15} & \text{4.71 $\pm$ 3.77} & \text{4.53 $\pm$ 3.48}  \\
$D^{\dagger}_{src}+D^{\dagger}_{s2t}$ & \multirow{2}{*}{VQGAN}  & \text{4.28 $\pm$ 3.66} & \text{4.51 $\pm$ 3.27} & \text{4.39 $\pm$ 3.47}     \\
$D_{src}+D_{s2t}$ & & \textbf{2.77 $\pm$ 2.38} & \textbf{3.51 $\pm$ 3.54} & \textbf{3.14 $\pm$ 3.04}  \\ \bottomrule
\end{tabular}
}

    \end{center}
\end{table}

\noindent\textbf{Ablations on Data Flow.}
Tab.~\ref{tab:ab_tol&s2t} presents the impact of different data generation strategies on model performance. We initiate our analysis by establishing a baseline using data simulated from an ideal lens model, which yields a relatively high MAE of 4.55 \si{\micro\meter} due to the domain gap. 
By replacing the ideal source with tolerance-aware simulation, the MAE decreases to 4.08 \si{\micro\meter}. This improvement demonstrates that incorporating tolerance perturbations during simulation effectively covers a wider range of lens samples, helping the model generalize better to real-world variations. 

We further evaluate the role of the domain transformation network $G$ in bridging the domain gap. 
While standard CycleGAN~\cite{zhu2017unpaired}-based enhancement shows limited gains, the VQGAN~\cite{esser2021taming}-based approach demonstrates superior performance. 
Specifically, the combination of tolerance-aware data and VQGAN-based enhancement achieves the lowest MAE of 3.14 \si{\micro\meter} in this comparison. 
This significant drop verifies that high-quality domain transformation is essential for generating pseudo-target images.

Crucially, even the best result here (3.14 \si{\micro\meter}) still lags behind our final DA3 performance (2.03 \si{\micro\meter}).
This gap indicates that data transformation alone is insufficient, motivating the subsequent integration of degradation-based augmentation and DA training.

\begin{table}[h!]
    \begin{center}
        \caption{\textbf{Ablation study on degradation-based augmentation types (Unit: \si{\micro\meter}).} 
        We compare the effect of applying four different types of degradation to the training data. 
        The baseline denotes the model trained on domain-transformation-enhanced data. Gaussian blur proves to be the most effective strategy.}
        \label{tab:ab_rd}
        \resizebox{0.5\textwidth}{!}{
\renewcommand{\arraystretch}{1.25}
\setlength{\tabcolsep}{1mm}
\begin{tabular}{@{}cccc@{}}
\toprule[0.17em]
Type of Degradation& 
\textbf{MAE$_{\text{X}}$ $\pm$ SD$_{\text{X}}$} &
\textbf{MAE$_{\text{Y}}$ $\pm$ SD$_{\text{Y}}$} &
\textbf{MAE$_{\text{Avg}}$ $\pm$ SD$_{\text{Avg}}$}  \\ \midrule
Baseline       & \text{2.41 $\pm$ 1.96} & \text{2.58 $\pm$ 2.47} & \text{2.49 $\pm$ 2.23}  \\
JPEG Compression    & \text{2.45 $\pm$ 2.15} & \text{2.89 $\pm$ 2.40} & \text{2.67 $\pm$ 2.29}  \\
\textbf{Gaussian Blur}  & \textbf{2.25 $\pm$ 1.86} & \textbf{2.15 $\pm$ 1.93} & \textbf{2.20 $\pm$ 1.90} \\
Gaussian Noise & \text{2.65 $\pm$ 1.97} & \text{2.49 $\pm$ 2.10} & \text{2.57 $\pm$ 2.04} \\
Random Masking & \text{2.17 $\pm$ 2.14} & \text{2.85 $\pm$ 2.53} & \text{2.51 $\pm$ 2.37} \\ \bottomrule
\end{tabular}
}

    \end{center}
\end{table}

\noindent\textbf{Ablations on Degradation-based Augmentation.}
To further bridge the domain gap, we investigate the impact of applying degradation-based data augmentation to the training data. We evaluate four degradation types detailed in Sec.~\ref{sec:DANN} as shown in Tab.~\ref{tab:ab_rd}. 

The Baseline row refers to the model trained with VQGAN-enhanced data but without additional augmentations, yielding an average MAE of 2.49 \si{\micro\meter}. 
However, adding degradations such as JPEG compression, Gaussian noise, or random masking degrades performance relative to the baseline. 
This suggests that high-frequency artifacts such as blockage or grain introduced by these methods may corrupt the precise edge features essential for regression.

In contrast, Gaussian blur significantly improves performance, reducing the average MAE to 2.20 \si{\micro\meter}. 
This improvement aligns with the real-world AA process, where unavoidable mechanical micro-vibrations often induce slight defocus or motion blur. By simulating this specific degradation pattern, the model becomes more robust to real-world imaging instabilities.
Consequently, we adopt the augmentation of Gaussian blur with random parameters as detailed in Sec.~\ref{exp:details} as the exclusive augmentation strategy for our final DA3 framework.

\noindent\textbf{Ablations on DA Training.}
Finally, we investigate the impact of the DA training strategy by performing a grid search over the hyperparameters for the adversarial loss $\lambda_{adv}$ and the pixel-wise consistency loss $\lambda_{pix}$.

\begin{figure}
  \centering
  \includegraphics[width=1.0\linewidth]{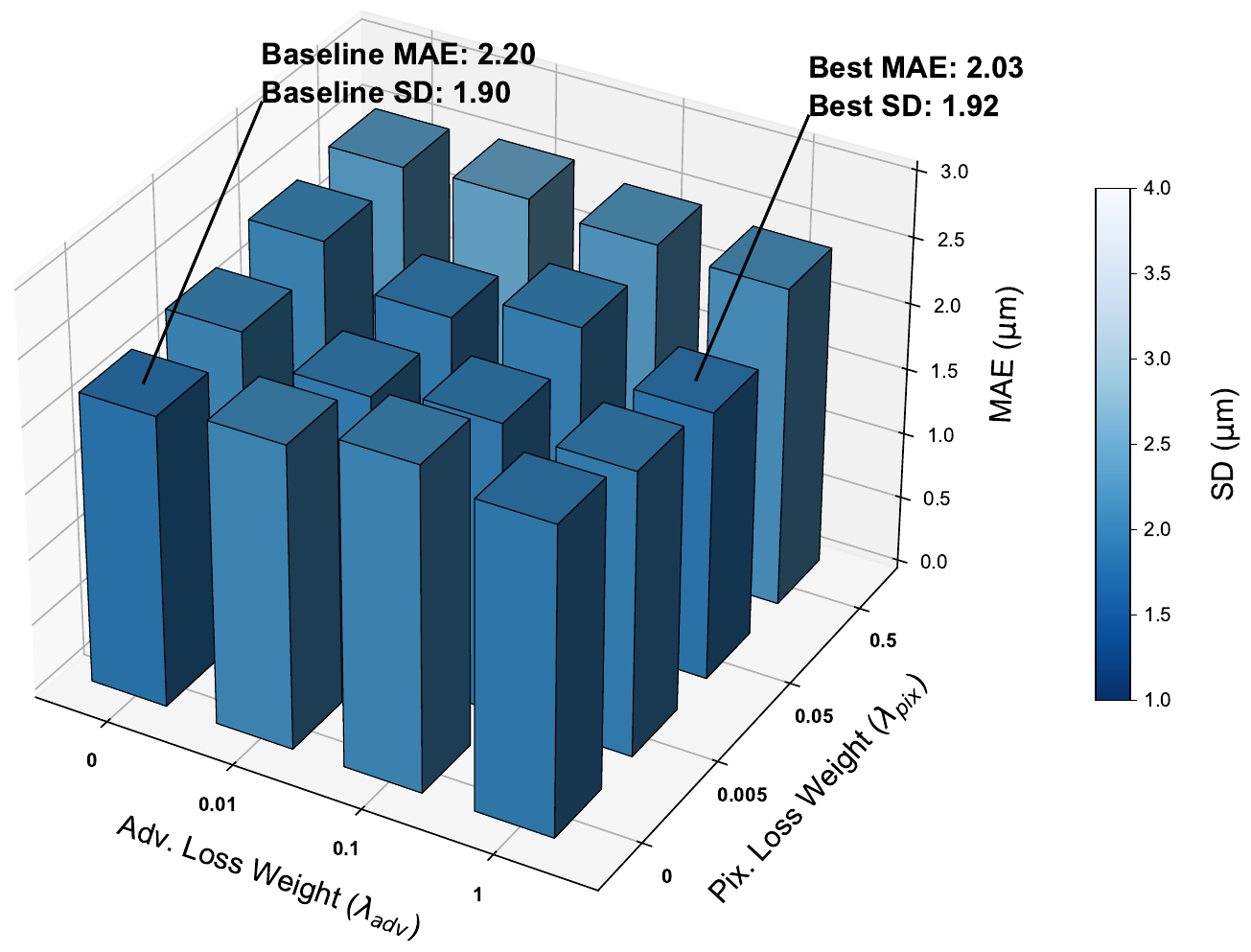}
  \caption{\textbf{Hyperparameter sensitivity analysis on DA training (Unit: \si{\micro\meter}).} 
  The 3D bar chart visualizes the MAE and SD on the security lens test set under different combinations of adversarial loss weight $\lambda_{adv}$ and pixel-wise loss weight $\lambda_{pix}$. 
  The height of the bars represents the MAE value, whereas the color intensity indicates the SD value.}
  \label{fig:paramstuning}
\end{figure}

As shown in Fig.~\ref{fig:paramstuning}, the configuration where both weights are zero ($\lambda_{adv}=0, \lambda_{pix}=0$) corresponds to the model trained solely with Gaussian blur augmentation, achieving a baseline MAE of 2.20 \si{\micro\meter}. 
By introducing the DA training objectives, the performance improves further. The optimal configuration is identified at $\lambda_{adv}=1$ and $\lambda_{pix}=0.05$, yielding the lowest MAE of 2.03 \si{\micro\meter}. 
This demonstrates that explicitly aligning feature distributions and enforcing pixel-level consistency provides an additional gain over data augmentation alone.
Furthermore, the 3D visualization reveals a broad effective region in the hyperparameter space.
Specifically, most combinations within the range of $\lambda_{adv} \in [0.01, 1]$ and $\lambda_{pix} \in [0.005, 0.05]$ consistently yield MAE values lower than the baseline (2.20 \si{\micro\meter}).
This indicates that the DA3 framework maintains stability within a reasonable parameter space, rather than relying on a narrow peak of optimal settings.
Such robustness eases the requirement for precise hyperparameter tuning, facilitating practical deployment in industrial scenarios.

\section{Conclusion and Discussion}
\label{sec:conclusion}
\subsection{Conclusion}
In this paper, we present DA3, a novel domain-adaptive framework designed to automate the AA process without relying on labor-intensive real-world data collection. 
By integrating tolerance-aware simulation, domain transformation, and domain adaptive training, our method effectively bridges the significant domain gap between simulated and real-world optical imaging systems. Experimental results demonstrate that DA3 not only generalizes well to lenses with varying tolerance combinations but also achieves robust and high-precision alignment comparable to upper-bound models trained on dense real-world datasets. 
This approach provides a scalable and cost-effective solution for mass lens production, significantly reducing the deployment cycle for new lens modules.

\subsection{Discussion and Future Work}
Despite the promising results, several aspects require further investigation to fully realize the potential of our framework.

First, our ablation studies indicate that the choice of the data transformation model has a substantial impact on the final regression performance. 
The performance advantage of VQGAN over CycleGAN suggests that the fidelity of the pseudo-target data is a critical bottleneck. 
Consequently, a primary focus of our future work involves exploring advanced data generation techniques. 
We plan to leverage emerging generative models, such as diffusion models~\cite{ho2020denoising,zhang2023adding,rombach2022high}, to achieve controllable, high-fidelity training data generation, thereby narrowing the initial domain gap at its source.

Second, while the current adversarial training strategy proves effective, the potential of DA training is far from exhausted. 
We plan to incorporate cutting-edge DA methodologies from other computer vision domains into our pipeline. 
Specifically, we aim to investigate mechanisms like contrastive learning~\cite{chen2020simple,kang2019contrastive} and disentangled representation learning~\cite{bousmalis2016domain,huang2018multimodal}. 
These approaches could enable the network to better decouple the misalignment-related degradation features from the domain-specific factors, leading to more robust generalization across diverse optical manufacturing scenarios.

\bibliographystyle{SELF-cas-model2-names}
\bibliography{ref}

@inproceedings{langehanenberg2015strategies,
  title={Strategies for active alignment of lenses},
  author={Langehanenberg, Patrik and Heinisch, Josef and Wilde, Chrisitan and Hahne, Felix and L{\"u}er{\ss}, Bernd},
  booktitle={Optifab 2015},
  volume={9633},
  pages={300--308},
  year={2015},
  organization={SPIE}
}

@inproceedings{ho2017precision,
  title={Precision lens assembly with alignment turning system},
  author={Ho, Cheng-Fang and Huang, Chien-Yao and Lin, Yi-Hao and Kuo, Hui-Jean and Kuo, Ching-Hsiang and Hsu, Wei-Yao and Chen, Fong-Zhi},
  booktitle={Optifab 2017},
  volume={10448},
  pages={429--434},
  year={2017},
  organization={SPIE}
}

@article{liu2024application,
  title={Application of deep learning in active alignment leads to high-efficiency and accurate camera lens assembly},
  author={Liu, Haibin and Li, Wenyong and Gao, Shaohua and Jiang, Qi and Sun, Lei and Zhang, Benhao and Zhao, Liefeng and Zhang, Jiahuang and Wang, Kaiwei},
  journal={Optics Express},
  volume={32},
  number={25},
  pages={43834--43849},
  year={2024},
  publisher={Optica Publishing Group}
}

@article{hu2025fast,
  title={Fast and accurate active alignment of camera lenses with physics-informed deep learning},
  author={Hu, Enjie and He, Jiajian and Zhou, Jingwen and Ren, Zheng and Sun, Huaze and Jiang, Tingting and Feng, Huajun and Chen, Yueting},
  journal={Optics Express},
  volume={33},
  number={10},
  pages={21256--21273},
  year={2025},
  publisher={Optica Publishing Group}
}

@article{burkhardt2025active,
  title={Active Alignments of Lens Systems with Reinforcement Learning},
  author={Burkhardt, Matthias and Schm{\"a}hling, Tobias and Stegmann, Pascal and Layh, Michael and Windisch, Tobias},
  journal={arXiv preprint arXiv:2503.02075},
  year={2025}
}

@article{wang2018deep,
  title={Deep visual domain adaptation: A survey},
  author={Wang, Mei and Deng, Weihong},
  journal={Neurocomputing},
  volume={312},
  pages={135--153},
  year={2018},
  publisher={Elsevier}
}

@inproceedings{long2015learning,
  title={Learning transferable features with deep adaptation networks},
  author={Long, Mingsheng and Cao, Yue and Wang, Jianmin and Jordan, Michael},
  booktitle={International conference on machine learning},
  pages={97--105},
  year={2015},
  organization={PMLR}
}

@article{tzeng2014deep,
  title={Deep domain confusion: Maximizing for domain invariance},
  author={Tzeng, Eric and Hoffman, Judy and Zhang, Ning and Saenko, Kate and Darrell, Trevor},
  journal={arXiv preprint arXiv:1412.3474},
  year={2014}
}

@inproceedings{ganin2015unsupervised,
  title={Unsupervised domain adaptation by backpropagation},
  author={Ganin, Yaroslav and Lempitsky, Victor},
  booktitle={International conference on machine learning},
  pages={1180--1189},
  year={2015},
  organization={PMLR}
}

@inproceedings{ghifary2016deep,
  title={Deep reconstruction-classification networks for unsupervised domain adaptation},
  author={Ghifary, Muhammad and Kleijn, W Bastiaan and Zhang, Mengjie and Balduzzi, David and Li, Wen},
  booktitle={European conference on computer vision},
  pages={597--613},
  year={2016},
  organization={Springer}
}

@inproceedings{zhu2017unpaired,
  title={Unpaired image-to-image translation using cycle-consistent adversarial networks},
  author={Zhu, Jun-Yan and Park, Taesung and Isola, Phillip and Efros, Alexei A},
  booktitle={Proceedings of the IEEE international conference on computer vision},
  pages={2223--2232},
  year={2017}
}

@article{jiang2025representing,
  title={Representing domain-mixing optical degradation for real-world Computational Aberration Correction via vector quantization},
  author={Jiang, Qi and Yi, Zhonghua and Gao, Shaohua and Gao, Yao and Qian, Xiaolong and Shi, Hao and Sun, Lei and Niu, Jinxing and Wang, Kaiwei and Yang, Kailun and others},
  journal={Optics \& Laser Technology},
  volume={183},
  pages={112080},
  year={2025},
  publisher={Elsevier}
}

@inproceedings{yang2021exploring,
  title={Exploring robustness of unsupervised domain adaptation in semantic segmentation},
  author={Yang, Jinyu and Li, Chunyuan and An, Weizhi and Ma, Hehuan and Guo, Yuzhi and Rong, Yu and Zhao, Peilin and Huang, Junzhou},
  booktitle={Proceedings of the IEEE/CVF International Conference on Computer Vision},
  pages={9194--9203},
  year={2021}
}

@article{li2024universal,
  title={A Universal Degradation-based Bridging Technique for Domain Adaptive Semantic Segmentation},
  author={Li, Wangkai and Sun, Rui and Zhang, Tianzhu},
  journal={arXiv preprint arXiv:2412.10339},
  year={2024}
}

@inproceedings{yue2019domain,
  title={Domain randomization and pyramid consistency: Simulation-to-real generalization without accessing target domain data},
  author={Yue, Xiangyu and Zhang, Yang and Zhao, Sicheng and Sangiovanni-Vincentelli, Alberto and Keutzer, Kurt and Gong, Boqing},
  booktitle={Proceedings of the IEEE/CVF international conference on computer vision},
  pages={2100--2110},
  year={2019}
}

@inproceedings{li2025towards,
  title={Towards unsupervised domain bridging via image degradation in semantic segmentation},
  author={Li, Wangkai and Sun, Rui and Mai, Huayu and Zhang, Tianzhu},
  booktitle={The Thirty-ninth Annual Conference on Neural Information Processing Systems},
  pages={1--32},
  year={2025}
}

@article{zhu2020deep,
  title={Deep subdomain adaptation network for image classification},
  author={Zhu, Yongchun and Zhuang, Fuzhen and Wang, Jindong and Ke, Guolin and Chen, Jingwu and Bian, Jiang and Xiong, Hui and He, Qing},
  journal={IEEE transactions on neural networks and learning systems},
  volume={32},
  number={4},
  pages={1713--1722},
  year={2020},
  publisher={IEEE}
}

@article{pandey2020target,
  title={Target-independent domain adaptation for WBC classification using generative latent search},
  author={Pandey, Prashant and Kyatham, Vinay and Mishra, Deepak and Dastidar, Tathagato Rai and others},
  journal={IEEE Transactions on Medical Imaging},
  volume={39},
  number={12},
  pages={3979--3991},
  year={2020},
  publisher={IEEE}
}

@inproceedings{sohn2017unsupervised,
  title={Unsupervised domain adaptation for face recognition in unlabeled videos},
  author={Sohn, Kihyuk and Liu, Sifei and Zhong, Guangyu and Yu, Xiang and Yang, Ming-Hsuan and Chandraker, Manmohan},
  booktitle={Proceedings of the IEEE International Conference on computer vision},
  pages={3210--3218},
  year={2017}
}

@article{liu2022blind,
  title={Blind image super-resolution: A survey and beyond},
  author={Liu, Anran and Liu, Yihao and Gu, Jinjin and Qiao, Yu and Dong, Chao},
  journal={IEEE transactions on pattern analysis and machine intelligence},
  volume={45},
  number={5},
  pages={5461--5480},
  year={2022},
  publisher={IEEE}
}

@inproceedings{chen2018domain,
  title={Domain adaptive faster r-cnn for object detection in the wild},
  author={Chen, Yuhua and Li, Wen and Sakaridis, Christos and Dai, Dengxin and Van Gool, Luc},
  booktitle={Proceedings of the IEEE conference on computer vision and pattern recognition},
  pages={3339--3348},
  year={2018}
}

@inproceedings{saito2019strong,
  title={Strong-weak distribution alignment for adaptive object detection},
  author={Saito, Kuniaki and Ushiku, Yoshitaka and Harada, Tatsuya and Saenko, Kate},
  booktitle={Proceedings of the IEEE/CVF conference on computer vision and pattern recognition},
  pages={6956--6965},
  year={2019}
}

@inproceedings{tsai2018learning,
  title={Learning to adapt structured output space for semantic segmentation},
  author={Tsai, Yi-Hsuan and Hung, Wei-Chih and Schulter, Samuel and Sohn, Kihyuk and Yang, Ming-Hsuan and Chandraker, Manmohan},
  booktitle={Proceedings of the IEEE conference on computer vision and pattern recognition},
  pages={7472--7481},
  year={2018}
}

@inproceedings{vu2019advent,
  title={Advent: Adversarial entropy minimization for domain adaptation in semantic segmentation},
  author={Vu, Tuan-Hung and Jain, Himalaya and Bucher, Maxime and Cord, Matthieu and P{\'e}rez, Patrick},
  booktitle={Proceedings of the IEEE/CVF conference on computer vision and pattern recognition},
  pages={2517--2526},
  year={2019}
}

@inproceedings{he2016deep,
  title={Deep residual learning for image recognition},
  author={He, Kaiming and Zhang, Xiangyu and Ren, Shaoqing and Sun, Jian},
  booktitle={Proceedings of the IEEE conference on computer vision and pattern recognition},
  pages={770--778},
  year={2016}
}

@inproceedings{wang2021unsupervised,
  title={Unsupervised real-world super-resolution: A domain adaptation perspective},
  author={Wang, Wei and Zhang, Haochen and Yuan, Zehuan and Wang, Changhu},
  booktitle={Proceedings of the IEEE/CVF international conference on computer vision},
  pages={4318--4327},
  year={2021}
}

@inproceedings{esser2021taming,
  title={Taming transformers for high-resolution image synthesis},
  author={Esser, Patrick and Rombach, Robin and Ommer, Bjorn},
  booktitle={Proceedings of the IEEE/CVF conference on computer vision and pattern recognition},
  pages={12873--12883},
  year={2021}
}

@article{Slor:26,
author = {Tomer Slor and Dean Oren and Shira Baneth and Tom Coen and Haim Suchowski},
journal = {Opt. Lett.},
number = {1},
pages = {169--172},
publisher = {Optica Publishing Group},
title = {Deep learning for optical misalignment diagnostics in multi-lens imaging systems},
volume = {51},
month = {Jan},
year = {2026}
}

@article{ho2020denoising,
  title={Denoising diffusion probabilistic models},
  author={Ho, Jonathan and Jain, Ajay and Abbeel, Pieter},
  journal={Advances in neural information processing systems},
  volume={33},
  pages={6840--6851},
  year={2020}
}

@inproceedings{zhang2023adding,
  title={Adding conditional control to text-to-image diffusion models},
  author={Zhang, Lvmin and Rao, Anyi and Agrawala, Maneesh},
  booktitle={Proceedings of the IEEE/CVF international conference on computer vision},
  pages={3836--3847},
  year={2023}
}

@inproceedings{rombach2022high,
  title={High-resolution image synthesis with latent diffusion models},
  author={Rombach, Robin and Blattmann, Andreas and Lorenz, Dominik and Esser, Patrick and Ommer, Bj{\"o}rn},
  booktitle={Proceedings of the IEEE/CVF conference on computer vision and pattern recognition},
  pages={10684--10695},
  year={2022}
}

@inproceedings{chen2020simple,
  title={A simple framework for contrastive learning of visual representations},
  author={Chen, Ting and Kornblith, Simon and Norouzi, Mohammad and Hinton, Geoffrey},
  booktitle={International conference on machine learning},
  pages={1597--1607},
  year={2020},
  organization={PmLR}
}

@inproceedings{kang2019contrastive,
  title={Contrastive adaptation network for unsupervised domain adaptation},
  author={Kang, Guoliang and Jiang, Lu and Yang, Yi and Hauptmann, Alexander G},
  booktitle={Proceedings of the IEEE/CVF conference on computer vision and pattern recognition},
  pages={4893--4902},
  year={2019}
}

@article{bousmalis2016domain,
  title={Domain separation networks},
  author={Bousmalis, Konstantinos and Trigeorgis, George and Silberman, Nathan and Krishnan, Dilip and Erhan, Dumitru},
  journal={Advances in neural information processing systems},
  volume={29},
  year={2016}
}

@inproceedings{huang2018multimodal,
  title={Multimodal unsupervised image-to-image translation},
  author={Huang, Xun and Liu, Ming-Yu and Belongie, Serge and Kautz, Jan},
  booktitle={Proceedings of the European conference on computer vision (ECCV)},
  pages={172--189},
  year={2018}
}

@article{ganin2016domain,
  title={Domain-adversarial training of neural networks},
  author={Ganin, Yaroslav and Ustinova, Evgeniya and Ajakan, Hana and Germain, Pascal and Larochelle, Hugo and Laviolette, Fran{\c{c}}ois and March, Mario and Lempitsky, Victor},
  journal={Journal of machine learning research},
  volume={17},
  number={59},
  pages={1--35},
  year={2016}
}

\section*{Funding}
This research was funded by the Natural Science Foundation of Zhejiang Province (Grant No. LZ24F050003), National Natural Science Foundation of China (Grant Nos. 12174341 and 62473139), Hunan Provincial Research and Development Project (Grant 2025QK3019), and State Key Laboratory of Autonomous Intelligent Unmanned Systems (the opening project number ZZKF2025-2-10).

\section{Data Generation and Preprocessing}
\label{sec:data}

\subsection{Simulation Data Generation}
We utilize the ZOS-API to automate the data generation process in Zemax. The procedure begins by optimizing the decenter parameters based on a predefined evaluation function to identify the position of optimal optical performance, which is marked as the origin $(0,0)$. Subsequently, we programmatically set a decenter sampling grid and acquire Point Spread Functions (PSFs) for 5 different Fields of View (FOVs) at each sampling position.

To synthesize the crosshair images, we start with a high-resolution base image ($90 \times 90$ pixels) containing an ideal cross pattern with a 1-pixel width. This base image undergoes an inverse Image Signal Processor (ISP) transformation to the RAW domain and is convolved with the spatially corresponding PSFs exported from Zemax. Finally, we apply a forward ISP pipeline, incorporating random gamma correction, scale jittering, and sensor noise simulation, to produce the final training samples. The images are then cropped to the target input sizes centered on the crosshair intersection.

\subsection{Real-world Data Preprocessing}
For the security lens setting, the raw images are captured at a resolution of $1440 \times 1080$. To isolate the region of interest, we first locate the crosshair center and crop a $280 \times 280$ region to remove redundant background information. This crop is then downsampled to $70 \times 70$ using bilinear interpolation to maintain structural consistency with the simulated data inputs.

\section{Detailed Network Architecture}
\label{sec:arch}

\subsection{Feature Extractor and Predictors}
The feature extractor $E$ is based on a ResNet-18 backbone. The final fully connected layer is replaced with an identity mapping to output a 512-dimensional feature vector. Both the label predictor $P$ and the domain classifier $D$ share a similar Multi-Layer Perceptron (MLP) architecture: a linear layer mapping 512 channels to 128 channels, followed by a ReLU activation and a dropout layer (rate=0.5). The predictor $P$ concludes with a linear layer outputting the 2D decenter value, while $D$ outputs a scalar logit.

\subsection{Domain Transformation Module}
The generator $G$ adopts a VQGAN-based U-Net structure with ResBlocks, utilizing a learned codebook to quantize deep features at the bottleneck. The discriminator $D_{s2t}$ follows a U-Net architecture equipped with spectral normalization to stabilize adversarial training.

\section{Training Strategy and Hyperparameters}
\label{sec:train}

\subsection{Training Settings}
The domain transformation generator $G$ is pre-trained for $20K$ iterations with a batch size of $16$ and a fixed learning rate of $1 \times 10^{-4}$. For the main DA3 training, we employ a step learning rate scheduler where the learning rate decays by a factor of 0.1 every 20 epochs. All models are implemented in PyTorch and trained on a single NVIDIA GeForce RTX 3090 GPU.

\subsection{Data Augmentation Details}
To bridge the domain gap and improve robustness against environmental disturbances, we apply several degradation-based augmentations. The specific configurations for these operations are as follows:

\begin{itemize}
    \item \textbf{JPEG compression ($\mathcal{T}_{jpeg}$):} Simulates compression artifacts and quality degradation with a scaling factor $q$ uniformly sampled from $\mathcal{U}(0.4, 0.7)$.
    \item \textbf{Gaussian blur ($\mathcal{T}_{blur}$):} Simulates optical defocus using a kernel size randomly selected from $\{3, 5, 7\}$ and a standard deviation $\sigma \sim \mathcal{U}(0.5, 2.0)$.
    \item \textbf{Gaussian noise ($\mathcal{T}_{noise}$):} Injects additive white Gaussian noise to simulate sensor noise, with a standard deviation $\delta \sim \mathcal{U}(0.02, 0.08)$.
    \item \textbf{Random masking ($\mathcal{T}_{mask}$):} Applies Cutout strategy to simulate partial occlusion or sensor defects, with an occlusion ratio $r \sim \mathcal{U}(0.05, 0.20)$.
\end{itemize}

\end{document}